\documentclass[conference]{IEEEtran}
\IEEEoverridecommandlockouts

\usepackage{cite}
\usepackage{amsmath,amssymb,amsfonts}
\usepackage{algorithmic}
\usepackage{graphicx}
\usepackage{textcomp}
\usepackage{xcolor}
\usepackage{amsmath}
\usepackage{bbm}
\usepackage{booktabs}
\usepackage{multirow}
\usepackage{fontawesome}
\usepackage{pifont} 
\usepackage{threeparttable}

\def\BibTeX{{\rm B\kern-.05em{\sc i\kern-.025em b}\kern-.08em
    T\kern-.1667em\lower.7ex\hbox{E}\kern-.125emX}}
\begin{document}

\title{Mirror-3DGS: Incorporating Mirror Reflections \\ into 3D Gaussian Splatting
}

\author{
	\IEEEauthorblockN{
		Jiarui Meng$^{1 \kern0.5pt \star}$, 
		Haijie Li$^{1 \kern0.5pt \star}$, 
		Yanmin Wu$^{1}$, 
		Qiankun Gao$^{1}$, 
		Shuzhou Yang$^{1}$,
        Jian Zhang$^{1 \kern0.5pt \dagger}$,
        Siwei Ma$^{2 \kern0.5pt \dagger}$
    }
    
	\IEEEauthorblockA{$^1$ School of Electronic and Computer Engineering, Peking University, China}
	\IEEEauthorblockA{$^2$ School of Computer Science, Peking University, China}
    \IEEEauthorblockA{\{mengjiarui, 2301212733, wuyanmin, gqk, szyang\}@stu.pku.edu.cn, \{zhangjian.sz, swma\}@pku.edu.cn}
}


\maketitle

\begin{abstract}
3D Gaussian Splatting (3DGS) has significantly advanced 3D scene reconstruction and novel view synthesis. However, like Neural Radiance Fields (NeRF), 3DGS struggles with accurately modeling physical reflections, particularly in mirrors, leading to incorrect reconstructions and inconsistent reflective properties. To address this challenge, we introduce \textbf{Mirror-3DGS}, a novel framework designed to accurately handle mirror geometries and reflections, thereby generating realistic mirror reflections. By incorporating mirror attributes into 3DGS and leveraging plane mirror imaging principles, Mirror-3DGS simulates a mirrored viewpoint from behind the mirror, enhancing the realism of scene renderings. Extensive evaluations on both synthetic and real-world scenes demonstrate that our method can render novel views with improved fidelity in real-time, surpassing the state-of-the-art Mirror-NeRF, especially in mirror regions.
\end{abstract}


\section{Introduction}
The 3D reconstruction and novel view synthesis are crucial in numerous applications, including the movie industry, computer gaming, virtual reality, and autonomous navigation. However, accurately capturing and rendering reflections from mirrors, which are prevalent in real-world scenes and significantly increase the visual complexity, remains a significant challenge for these technologies.

\renewcommand{\thefootnote}{}
\footnotetext{$^{\star}$ means equal contribution and $^{\dagger}$ means corresponding authors. This work was supported in part by the National Natural Science Foundation of China no. 61931014, U21B2012, in part by the Beijing Natural Science Foundation no. L242014, and in part by the Fundamental Research Funds for the Central Universities.}


Traditional methods~\cite{srinivasan2021nerv, verbin2022ref, zhang2022modeling, shi2023gir, ma2023specnerf} simulate surface reflections by decomposing material properties and lighting conditions to handle scenes with mirrors. However, they often struggle to accurately infer surfaces and mirror components, especially in managing pure mirror reflections. With advancements in Neural Radiance Fields (NeRF), new methods~\cite{verbin2022ref, guo2022nerfren, zeng2023mirror} have emerged. NeRFReN~\cite{guo2022nerfren} enhances rendering quality by using independent radiance fields for reflective and transmissive parts. Mirror-NeRF~\cite{zeng2023mirror} employs Whitted Ray Tracing~\cite{whitted2005improved} for physically accurate mirror reflections, achieving high realism in novel view synthesis. Despite these improvements, NeRF-based methods face challenges such as long training times and slow rendering speeds, limiting their real-time application.
\begin{figure}[t]
\centering
\includegraphics[width=\columnwidth]{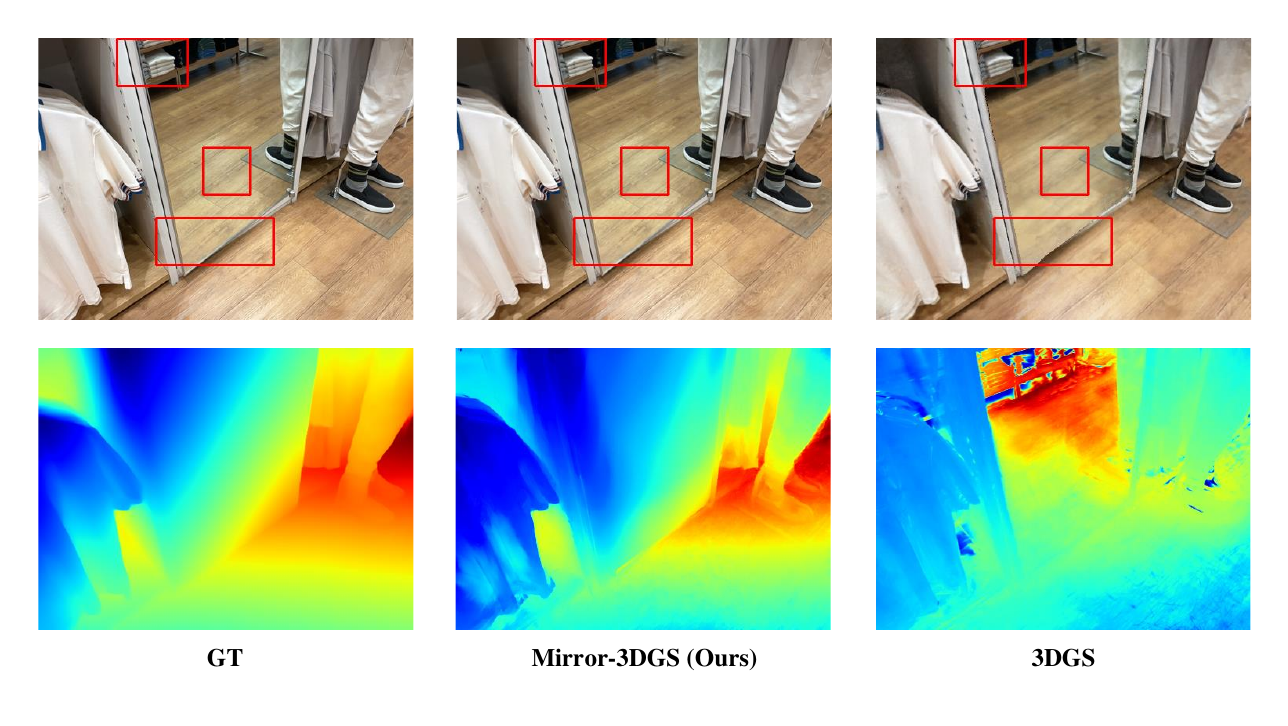}
\caption{
\textbf{The synthesized novel views and depth maps.} 3DGS confuses mirror reflection, resulting in erroneous mirror depth. Ours properly handles content in the mirror with depth map almost identical to GT.
}
\vspace{-4mm}
\label{figure: depth}
\end{figure}

Recently, 3D Gaussian Splatting (3DGS)~\cite{3dgs} has emerged as a promising approach for real-time, high-fidelity 3D scene rendering. Instead of the slow volumetric rendering of NeRF, 3DGS explicitly represents scenes using 3D Gaussians and rasterizes 3D Gaussians onto the image plane using splatting techniques, achieving significant improvements in training and rendering speeds~\cite{yang2024fourier123}. However, 3DGS tends to treat mirror reflections as separate scenes, constructing Gaussian points for them as if they physically exist, 
leading to inaccurate capture of these virtual images within mirrors during reconstruction and rendering, affecting the realism and visual coherence of the scene, as shown in  \textcolor{red}{Fig.~\ref{figure: depth}}.

To address these challenges, we propose Mirror-3DGS, a novel rendering framework based on 3DGS, aimed at producing high-fidelity novel views in mirror-containing scenes. \textit{Our Mirror-3DGS is centered on the principle that the virtual image presented on the mirror is equivalent to observing the real world from behind the mirror.} In general, We first identify 3D Gaussians representing the mirror, then observe from both the current and mirrored viewpoints, integrating these perspectives to render the final image. We introduce an attribute to signify mirror-like properties to 3D Gaussians and estimate the mirror plane equation, allowing us to derive the mirror transformation matrix and mirrored camera parameters. Our method involves a two-stage training process: 1) During the first stage, We learn a rough 3D Gaussian representation of the scene without mirror content and estimate the mirror plane equation. The mirror content is replaced with a fixed color based on the ground truth mirror mask to better learn mirror attributes and avoid interference from virtual images.
2) In the second stage, We filter Gaussians with high mirror properties for precise estimation of the mirror equation. By integrating views from both the current and mirrored viewpoints, we optimize the rendering quality of the entire scene.To ensure stability and improve the accuracy of plane estimation, we introduce planar consistency and depth constraints throughout the training process. These constraints help achieve more realistic and refined rendering results. Experiments on three synthetic and three real-world scenes demonstrate that Mirror-3DGS delivers rendering quality comparable to the state-of-the-art Mirror-NeRF, while significantly reducing training time and enabling real-time rendering. Our contributions can be summarized as follows:
\begin{itemize}
    \item \textbf{Innovative Rendering Framework for Mirror Scenes}. We present Mirror-3DGS, a novel rendering framework that addresses the limitations of 3DGS in handling mirror scenes, enabling realistic synthesis of novel views.
    
    \item \textbf{Precise Physical Modeling of Mirror Reflection}. We leverage the principle of plane mirror imaging to enhance 3DGS, enabling accurate modeling of physical space and producing high-fidelity rendering results.

    \item \textbf{Extensive Experiments}. We conduct comprehensive experiments across synthetic and real scenes to validate our proposed method. The results demonstrate our superiority in real-time rendering of high-quality novel views, particularly surpassing the state-of-the-art in mirror regions.
\end{itemize}

\section{Methodology}
\subsection{Preliminaries}

3D Gaussian Splatting (3DGS) is a state-of-the-art approach in novel view synthesis, utilizing the splatting technique~\cite{yifan2019differentiable} for efficient real-time rendering. In 3DGS, the scene is represented using anisotropic Gaussians characterized by center positions $\boldsymbol{\mu} \in \mathbb{R}^3$, covariances $\mathbf{\Sigma} \in \mathbb{R}^{3\times3}$, color defined by spherical harmonic (SH) coefficients $\boldsymbol{c} \in \mathbb{R}^{3\times (k+1)^2}$ (with $k$ denoting the SH order), and opacity $\alpha \in \mathbb{R}^1$. The 3D Gaussian can be queried as follows:
\begin{equation}
	\label{eq: gaussian}
	G(\mathbf{x})=e^{-\frac{1}{2}(\mathbf{x}-\boldsymbol{\mu})^{\top}\mathbf{\Sigma}^{-1}(\mathbf{x}-\boldsymbol{\mu})},
\end{equation}
where $\mathbf{x}$ represents the position of the query point. Subsequently, an efficient 3D to 2D Gaussian mapping~\cite{zwicker2001ewa} is employed to project the Gaussian onto the image plane:
\begin{equation}
	\boldsymbol{\mu}^{\prime}={\mathbf{P}\mathbf{W}\boldsymbol{\mu}}, 
\quad
	\mathbf{{\Sigma}^{\prime}}=\mathbf{J}\mathbf{W} \mathbf{\Sigma} \mathbf{W^{\top}}\mathbf{J^{\top}},
\end{equation}
where $\boldsymbol{\mu}^{\prime}$ and $\mathbf{{\Sigma}^{\prime}}$ separately represent the 2D mean position and covariance of the projected 3D Gaussian. $\mathbf{P}$, $\mathbf{W}$ and $\mathbf{J}$ denote the projective transformation, viewing transformation, and Jacobian of the affine approximation of $\mathbf{P}$, respectively. 
The color of the pixel on the image plane, denoted by $\mathbf{p}$, uses a typical neural point-based rendering~\cite{kopanas2022neural,KPLD21}. Let  $\mathbf{C} \in \mathbb{R}^{H\times W\times3}$ represent the color of rendered image where $H$ and $W$ represents the height and width of images, the rendering process outlined as follows:
\begin{equation}
\begin{split}
    \sigma_i=\alpha_ie^{-\frac12(\mathbf{p}-\boldsymbol{\mu}^{\prime})^{\top}\mathbf{{{\Sigma}^{\prime}}^{-1}}(\mathbf{p}-\boldsymbol{\mu}^{\prime})}.
    \\
	{\mathbf{C}(\mathbf{p})} = {\sum_{i=1}^{N}
	c_{i}\sigma_{i}
	\prod_{j=1}^{i-1}(1-\sigma_{j})},
\label{eq_render}
\end{split}
\end{equation}
where $N$ represents the number of sample Gaussians that overlap the pixel $\mathbf{p}$,   $c_{i}$ and $\alpha_{i}$ denote the color and opacity of the i-th Gaussian, respectively.

\begin{figure}[t]
\centering
\includegraphics[width=\columnwidth]{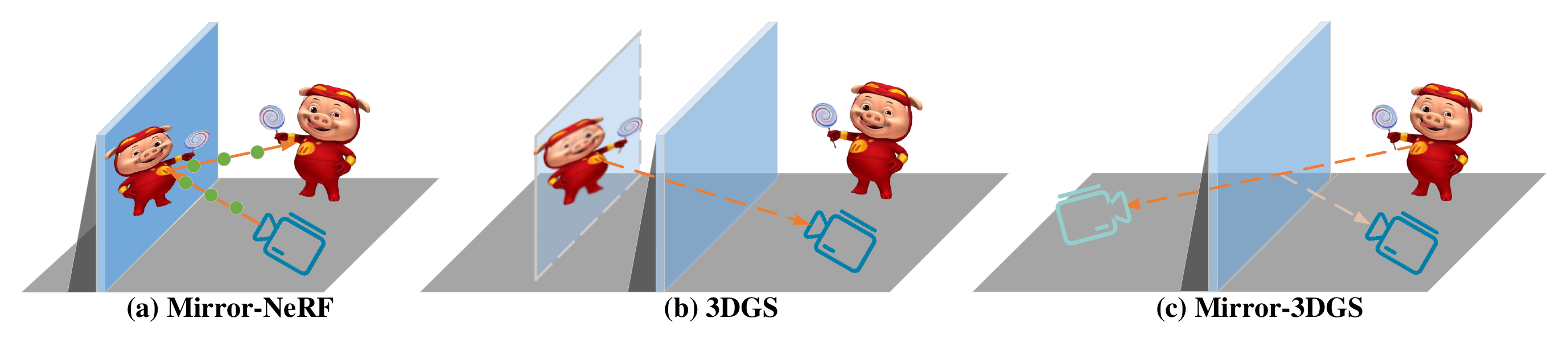}
\caption{\textbf{The underlying principles in handling mirrored contents}. \textbf{a)} Mirror-NeRF employs a technique involving ray tracing and sampling to achieve mirror reflection. \textbf{b)} 3DGS mistakenly considers objects reflected by the mirror to be placed behind it, resulting the floaters behind the mirror.  \textbf{c)} Our proposed Mirror-3DGS, captures the reflected objects by simulating a novel viewpoint situated behind the mirror.} 
\vskip -0.2in
\label{figure: GGBOND}
\end{figure}

\subsection{Mirror-aware 3D Gaussian Representation}
\label{sec:3.2}
In the real world, a point's color varies with different viewpoints due to its material and environmental lighting. While NeRF and 3DGS use view-dependent MLPs and spherical harmonics to simulate this, mirrors' extreme specularity amplifies view-dependent inconsistencies. We review existing solutions and introduce our reflection-agnostic Mirror-3DGS.

\textbf{$\bullet$ Mirror-NeRF}: Mirror-NeRF achieves mirror modeling by learning a unified probabilistic reflectance field and simulates mirror reflection through physics-based ray tracing, enabling accurate mirror rendering, as illustrated in \textcolor{red}{Fig.~\ref{figure: GGBOND}(a)}. However, its limitations lie in the intensive computational cost of ray tracing and the prolonged NeRF training time.

\textbf{$\bullet$  3DGS}: 3DGS is based on explicit gaussian spheres representation and alpha-blending along ray directions, tending to generate points behind mirrors as shown in \textcolor{red}{Fig.~\ref{figure: GGBOND}(b)}. This violates physical laws and undermines the rendering quality of mirror back-side scenes.


\textbf{$\bullet$  Mirror-3DGS (Ours)}: We innovatively introduce a mirror imaging method -- ``\textbf{moving behind the mirror for a view of the world}'', where the virtual image presented on the mirror is equivalent to observing the real world from behind the mirror, as illustrated in \textcolor{red}{Fig.~\ref{figure: GGBOND}(c)}. Therefore, we avoid the time-consuming ray tracing required by Mirror-NeRF. 

More specifically, \textbf{1)} we render an image from the original viewpoint, \textit{i.e.} observing from in front of the mirror, and retain the parts outside the mirror; \textbf{2)} we mirror the original viewpoint to behind the mirror, rendering an image from the virtual viewpoint, and only keep the parts of the mirror (see \textcolor{red}{Sec.~\ref{sec:virtual_view}}); \textbf{3)} finally, we merge these two images to form the final result (see \textcolor{red}{Sec.~\ref{sec:virtual_view}}).

To realize the first step above, we need to enable mirror detection. We add a learnable mirror attribute $m$ $\in$ $[0, 1]$ for each Gaussian in the original 3DGS, representing the probability of it being a mirror. The rendering process is similar to color rendering:
\begin{equation}
    {\mathbf{M}(\mathbf{p})}=\sum_{i=0}^{N} m_i \sigma_i\prod_{j=1}^{i-1}\left(1-\sigma_j\right),
    \label{eq:mask_render}
\end{equation}
where  $\mathbf{M}\in \mathbb{R}^{H\times W\times 1}$ represents the rendered mirror mask, and the rest of the symbols are defined as in \textcolor{red}{Eq.~\eqref{eq_render}}. We utilize the GT mirror mask $\mathbf{M}_{gt}$ provided by the dataset to supervise the learning of the mirror attribute:
\begin{equation}
    \mathcal{L}_{mask} = \mathcal{L}_1(\mathbf{M}, \mathbf{M}_{gt}).
    \label{eq:mask_loss}
\end{equation}

\subsection{Virtual Mirrored Viewpoint Construction and Image Fusion}
\label{sec:virtual_view}

\textbf{Mirror Parameterization.} In the previous section, we acquire the mirror attribute for each Gaussian. Using this attribute and opacity, we filter out Gaussians belonging to the mirror to construct a plane in 3D space. We adopt the parameterization of an infinite plane~\cite{kaess2015simultaneous}: $\pi = (\boldsymbol{n}_{\pi}^\top , d)$ , where $\boldsymbol{n}_{\pi}= (a,b,c) ^{\top}$ represents the normal of the plane, and $\frac{d}{\left \| \boldsymbol{n}_{\pi}  \right \| } $ represents the distance from the origin of the plane to the world coordinate origin. Thereby, any point $\boldsymbol{p}  =(x,y,z)^{\top}$ on the plane satisfies: 
\begin{equation}
    \boldsymbol{n}_{\pi}^\top \boldsymbol{p} + d = 0.
\end{equation}
The optimization of the plane $\pi$ is presented in \textcolor{red}{Sec.~\ref{sec:traing}}.

\textbf{Virtual Mirrored Viewpoint Construction.} Based on the above plane equation, given an original observation viewpoint $\mathbf{P_o} \in \mathbb{R} ^{4\times 4}$, our goal is to obtain the mirrored viewpoint $\mathbf{P_m} \in \mathbb{R} ^{4\times 4}$ of this viewpoint relative to the plane. This transformation can be formulated as: 
\begin{equation}
\label{eq: trans}
    \mathbf{P_m} = \mathbf{T_m} \mathbf{P_o},
\end{equation}
where we define $\mathbf{T_m}$ as the mirror transformation matrix, represented as:
\begin{equation}
    \mathbf{T_m} =\left[\begin{matrix}1-2a^2&-2ab&-2ac&-2ad\\-2ab&1-2b^2&-2bc&-2bd\\-2ac&-2bc&1-2c^2&-2cd\\0&0&0&1\\\end{matrix}\right].
\end{equation}

\textbf{Image Fusion.} The image from the mirror viewpoint is obtained using the same rendering equation. We acquire the rendered image $\mathbf{C}_{\mathbf{P_o}} \in \mathbb{R}^{H\times W\times3}$ from the original viewpoint, the rendered image $\mathbf{C}_{\mathbf{P_m}} \in \mathbb{R}^{H\times W\times3}$ from the mirror viewpoint, and the mirror mask $\mathbf{M}$. The final composite image is fused as follows:
\begin{equation}
    \mathbf{C}_{fuse}= \mathbf{C}_{\mathbf{P_o}} \odot  {(\mathbbm{1}-\mathbf{M}})+ \mathbf{C}_{\mathbf{P_m}} \odot \mathbf{M},
    \label{eq:fuse_image}
\end{equation}
where $\odot$ means element-wise multiplication.

\subsection{Two-Stage Training Strategy}
\label{sec:traing}

\begin{figure}[t]
\centering
\includegraphics[width=\columnwidth]{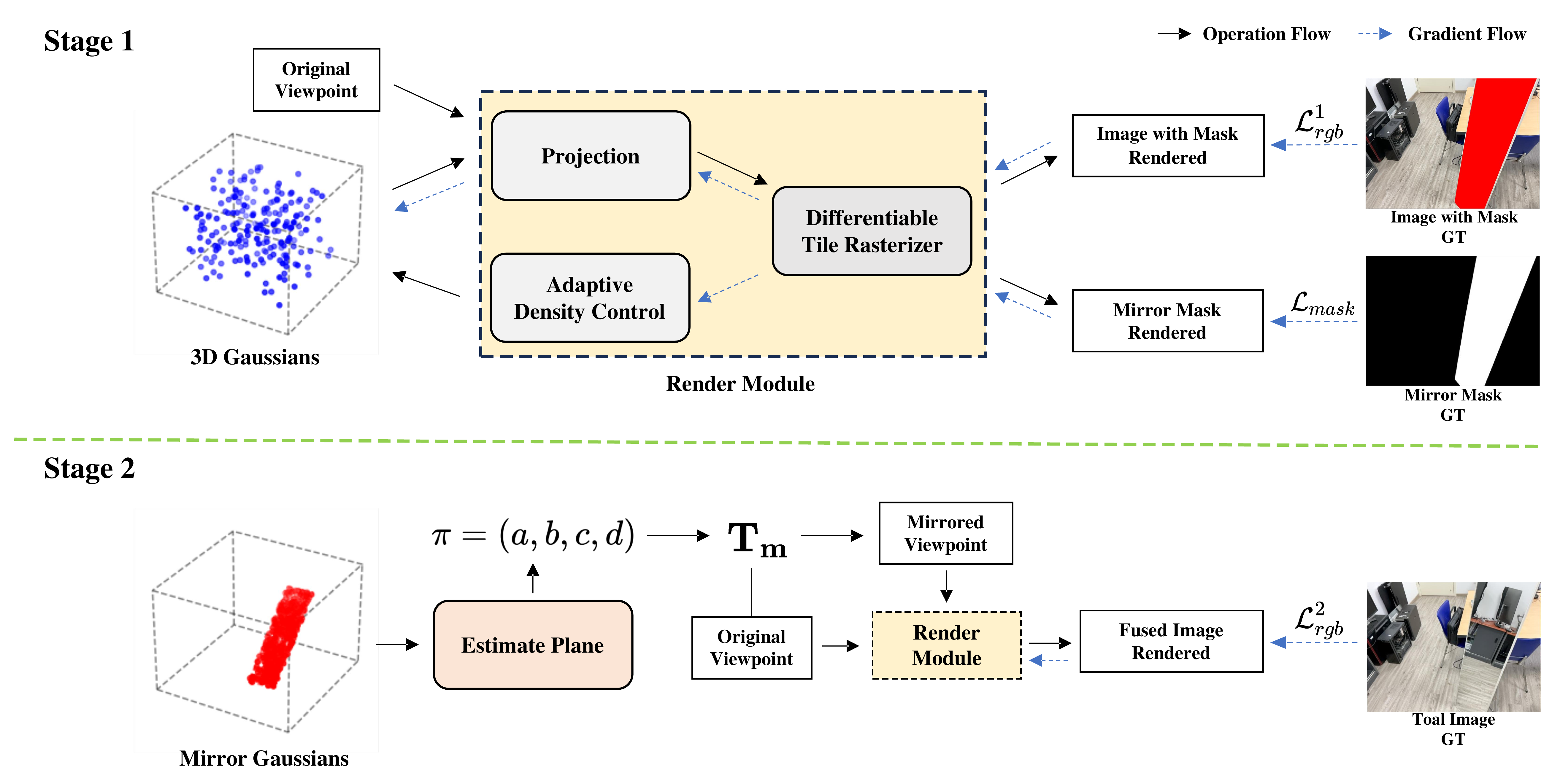}
\caption{\textbf{The illustration of our two-stage training pipeline.} In the \textbf{Stage 1}, we use the mirror mask and view image with mirror content replaced by red to learn the mirror plane and a coarse 3D Gaussian representation of the scene.  In the \textbf{Stage 2},based on the estimated mirror plane, we fuse views from the original and mirrored viewpoints to form the final rendered result, further optimizing the 3D Gaussians.} 
\label{figure:train pipe}
\vspace{-5mm}
\end{figure}

We use a two-stage training strategy, as illustrated in \textcolor{red}{Fig.~\ref{figure:train pipe}}.

\textbf{Stage 1:} The training objective of the first stage is to obtain a rough Gaussian representation without mirror content and to optimize the mirror equation. We use the GT mirror mask provided by the dataset to erase the mirror content in the GT image  $\mathbf{C}_{gt}$ and fill it with red, denoted as $\mathbf{C}_{wo}^{m}$. This is done to optimize the mirror attribute of Gaussians and to avoid generating false-Gaussians inside the mirror. The loss for rendering RGB images is consistent with 3DGS\cite{3dgs}, \textit{i.e.}, L1 loss and D-SSIM term: 
\begin{equation}
    \mathcal{L}_{rgb}^1 = (1 \! - \! \gamma ) \mathcal{L}_1(\mathbf{C}_{\mathbf{P_o}}, \mathbf{C}_{wo}^{m})  +  \gamma\mathcal{L}_{D-SSIM}(\mathbf{C}_{\mathbf{P_o}}, \mathbf{C}_{wo}^{m})
\end{equation}
where $\gamma$ is the weight, and $\mathbf{C}_{\mathbf{P_o}}$ is the rendered image from the original viewpoint, as defined in \textcolor{red}{Sec.~\ref{sec:virtual_view}}.

Additionally, to improve the quality of rendering and geometric structure, we add depth supervision. First, a pretrained depth estimation model~\cite{costanzino2023learning} customized for mirror scenes is utilized to extract the depth map $\mathbf{D}_{gt}$ from image $\mathbf{C}_{wo}^{m}$.  Then a modified rasterization library~\cite{luiten2024dynamic} is used to render the predicted depth map $\mathbf{D}_{pred}$,  supervised with a similar loss: 
\begin{equation}
\label{eq:depth}
    \mathcal{L}_{depth} = \mathcal{L}_1(\mathbf{D}_{pred}, \mathbf{D}_{gt}) .
\end{equation}

The first stage focuses on optimizing the mirror equation $\pi = (\boldsymbol{n}_{\pi}^\top, d)$ to accurately generate virtual mirrored viewpoints. We filter the Gaussians with high mirror probability and use an improved RANSAC algorithm~\cite{fischler1981random} to remove noise Gaussians. Then we supervise that the remaining Gaussians fall on the mirror plane: 
\begin{equation}
    \mathcal{L}_{plane}=\frac{1}{N_f} \sum_{i=1}^{N_f}\left| \boldsymbol{n}_{\pi}^\top  \boldsymbol{p}_i + d \right|,
\end{equation}
where $\boldsymbol{p}_i$ is the coordinate of the Gaussian to be optimized, $N_f$ is the number of filtered mirror Gaussians. The overall loss for the first stage is: 
\begin{equation}
    \mathcal{L}_{stage}^{1} = \lambda_{mask}\mathcal{L}_{mask} + \mathcal{L}_{rgb}^{1} + \lambda_{depth}\mathcal{L}_{depth} + \mathcal{L}_{plane},
\end{equation}
where $\lambda_{mask}, \lambda_{depth}$ are the weight of mask loss and depth loss, respectively.

\textbf{Stage 2:} The objective of the second stage is to refine the rendering quality of the fused images, during which the mirror equation is fixed. As described in \textcolor{red}{Sec.~\ref{sec:virtual_view}}, we fuse images from the current viewpoint and the virtual mirrored viewpoint, supervising with the complete, unmasked GT image $\mathbf{C}_{gt}$: 
\begin{equation}
\label{eq:loss_rgb_2}
    \mathcal{L}_{rgb}^{2} = (1 \!- \!\gamma ) \mathcal{L}_1(\mathbf{C}_{fuse}, \mathbf{C}_{gt}) + \gamma\mathcal{L}_{D-SSIM}(\mathbf{C}_{fuse}, \mathbf{C}_{gt})
\end{equation}
where $\mathbf{C}_{fuse}$ is the fused rendered image (see \textcolor{red}{Eq.~\eqref{eq:fuse_image}}). The overall loss of the second stage is expressed as: 
\begin{equation}
    \mathcal{L}_{\text {stage}}^{2} = \lambda_{mask}\mathcal{L}_{mask} + \mathcal{L}_{rgb}^{2}.
\end{equation}

\begin{figure*}[ht]
\centering
\includegraphics[width=\textwidth]{file/Qualitative_NEW.pdf}
\caption{\textbf{Qualitative comparison of novel view synthesis.} Two raws display washroom and discussion room scenes. The top-right shows magnified mirror regions, and the bottom-right shows predicted mirror masks. PSNR is calculated only for the mirror regions. Both Mirror-NeRF and Mirror-3DGS accurately predict mirror masks, with Mirror-3DGS achieving the best performance.
} 
\label{figure:qualitative_comparison}
\vskip -0.2in
\end{figure*}
\setlength{\tabcolsep}{0.1pt}

\begin{table}[tb]
  \caption{Quantitative comparison on the Mirror-NeRF dataset. Best results are \textbf{bolded} and second best are \underline{underlined}. Methods marked with `*' are evaluated only on mirror regions.}
  \label{table:complete_test_comparison}
  \centering
  \scriptsize
  \begin{tabular}{lccccccccc}
    \toprule
    \multirow{2}{*}{Method} & \multicolumn{4}{c}{Synthetic} && \multicolumn{4}{c}{Real} \\
    \cmidrule{2-5} \cmidrule{7-10}
    & PSNR↑ & SSIM↑ & LPIPS↓ & FPS↑ && PSNR↑ & SSIM↑ & LPIPS↓ & FPS↑ \\
    \cmidrule{2-5} \cmidrule{7-10}
    InstantNGP~\cite{muller2022instant} & 23.54 & 0.71 & 0.42 & 34.89 && 10.51 & 0.20 & 0.71 & 34.36 \\
    DVGO~\cite{sun2022direct} & 28.05 & 0.82 & 0.29 & 4.35 && 22.18 & 0.67 & 0.33 & 4.46 \\
    Mirror-NeRF~\cite{zeng2023mirror} & \textbf{38.07} & \textbf{0.99} & \textbf{0.01} & 0.75 && \underline{25.04} & \textbf{0.86} & \textbf{0.06} & 0.07 \\
    \midrule
    3DGS~\cite{3dgs} & 37.00 & \underline{0.97} & 0.04 & \textbf{341.13} && 22.55 & 0.73 & 0.24 & \textbf{209.62} \\
    \textbf{Mirror-3DGS (Ours)} & \underline{37.89} & \underline{0.97} & \textbf{0.01} & \underline{171.98} && \textbf{25.93} & \underline{0.81} & \textbf{0.06} & \underline{95.97} \\
    \midrule
    3DGS*~\cite{3dgs} & 28.35 & - & - & - && 21.99 & - & - & - \\
    Mirror-NeRF*~\cite{zeng2023mirror} & 29.47 & - & - & - && 22.39 & - & - & - \\
    \textbf{Mirror-3DGS* (Ours)} & \textcolor{blue}{\textbf{32.40}} & - & - & - && \textcolor{blue}{\textbf{24.53}} & - & - & - \\
    \bottomrule
  \end{tabular}
\vspace{-3mm}
\end{table}

\section{Experiments}
\label{sec:blind}

\subsection{Experimental Setup}

\noindent\textbf{Dataset}.
We utilize the publicly released dataset by Mirror-NeRF~\cite{zeng2023mirror}, which includes common scenarios with prevalent mirrors. The dataset comprises three synthetic scenes (living room, office, and washroom) and three real-world scenes (discussion room, lounge, and market). Each scene has 200 to 300 training views with accompanying mirror masks. 

\noindent\textbf{Implementation}. Training and testing are done at $480\times360$ resolution for real scenes and $400\times400$ for synthetic scenes. The training process is divided into two stages:  we train 5000 steps in stage 1, followed by 65000 steps in stage 2. The loss weights $\lambda_{mask}$ and $\lambda_{depth}$ are set to 1 and 0.1, respectively.

\noindent\textbf{Evaluation}. 
The rendering quality of all methods is evaluated and compared using the widely recognized PSNR,  SSIM~\cite{ssim}, and LPIPS~\cite{lpips} metrics.
Additionally, we compare the rendering speed using FPS (Frames Per Second). All experiments are conducted on a single RTX 3090 GPU for fair comparison. 

\subsection{Experimental Results}

\noindent\textbf{Quantitative Comparisons}.
We conduct quantitative comparisons between our method and several competitive counterparts, and
the comparison results are detailed in \textcolor{red}{Tab.~\ref{table:complete_test_comparison}}.

Among these competitors, Mirror-NeRF achieves high PSNR, SSIM, and LPIPS scores in synthetic and real scenes. However, its rendering speed is significantly lower due to NeRF's volumetric rendering inefficiencies. Even the fastest NeRF-based method, InstantNGP, is much slower than 3DGS and our approach in real scenes. Therefore, NeRF-based methods are challenging for real-time rendering scenarios.

Our method yields performance better than vanilla 3DGS. \textit{Specifically, our Mirror-3DGS demonstrates an average PSNR improvement of 0.89 and 3.38 dB over vanilla 3DGS on the synthetic and real scenes, respectively.} Additionally, our SSIM and LPIPS metrics far exceed those of vanilla 3DGS, indicating that our synthesized views are significantly more realistic. These improvements are mainly attributed to our method's specialized handling of mirrors and their reflections in the scene, which inevitably introduces some additional computation overhead. Therefore, while our rendering speed is slightly lower than that of vanilla 3DGS, it is significantly faster than NeRF-based methods, making it suitable for real-time rendering applications. 

It is important to highlight that \textit{our Mirror-3DGS outperforms both Mirror-NeRF and vanilla 3DGS in terms of the PSNR metric when focusing solely on the mirror regions}, as shown in the last three rows of \textcolor{red}{Tab.~\ref{table:complete_test_comparison}}, demonstrating our method's enhanced capability in handling mirror reflections.

\noindent\textbf{Qualitative Comparisons}.
The qualitative analysis presented in
\textcolor{red}{Fig.~\ref{figure:qualitative_comparison}}, focusing on the detailed rendering of mirrors.  Our Mirror-3DGS, by incorporating the principles of plane mirror imaging and accurately modeling spatial geometric relationships, enables high-fidelity rendering of mirror contents. Across all six scenes, our Mirror-3DGS achieves rendering quality comparable to Mirror-NeRF, demonstrating its effectiveness in handling complex mirror reflection.

\section{Conclusion}
In this paper, we introduce Mirror-3DGS, a novel rendering framework that addresses the challenge of accurately capturing and rendering physical reflections in mirror-containing scenes. Our Mirror-3DGS employs a two-stage training process that first filters mirror Gaussians and then uses this information to accurately estimate the mirror plane and derive mirrored camera parameters. The effectiveness of Mirror-3DGS in delivering high-quality renderings of scenes with mirrors is demonstrated by experiments on 
 synthetic and real scenes. 
 


\bibliographystyle{IEEEtran}
\bibliography{references}

\end{document}